\documentclass[a4paper, 10pt, conference]{ieeeconf}      

\IEEEoverridecommandlockouts                              

\makeatletter  
\newif\if@restonecol  
\makeatother

\usepackage[linesnumbered,ruled,vlined]{algorithm2e}
\usepackage{algpseudocode}  
\usepackage{amsmath}  

\usepackage{graphicx}
\usepackage{subfigure}
\usepackage{amssymb}

\title{\LARGE \bf
Multi-Tree Guided Efficient Robot Motion Planning
}

\author{Zhirui Sun, Jiankun Wang, \emph{Senior Member, IEEE} and Max Q.-H. Meng, \emph{Fwllow, IEEE} 
\thanks{This work is partially supported by Shenzhen Key Laboratory of Robotics Perception and Intelligence (ZDSYS20200810171800001), Southern University of Science and Technology, Shenzhen 518055, China, \emph{(Corresponding authors:Jiankun Wang, Max Q.-H. Meng).}}
\thanks{Zhirui Sun and Jiankun Wang are with Shenzhen Key Laboratory of Robotics Perception and Intelligence, and the Department of Electronic and Electrical Engineering, Southern University of Science and Technology, Shenzhen 518055, China. {\tt\small e-mail: wangjk@sustech.edu.cn}}%
\thanks{Max Q.-H. Meng is with Shenzhen Key Laboratory of Robotics Perception and Intelligence, and the Department of Electronic and Electrical Engineering, Southern University of Science and Technology, Shenzhen 518055, China, on leave from the Department of Electronic Engineering, The Chinese University of Hong Kong, Hong Kong, and also with the Shenzhen Research Institute, The Chinese University of Hong Kong in Shenzhen, Shenzhen 518057, China. {\tt\small e-mail: max.meng@ieee.org}}%
}

\begin{document}

\maketitle
\thispagestyle{empty}
\pagestyle{empty}

\begin{abstract}

Motion Planning is necessary for robots to complete different tasks. Rapidly-exploring Random Tree (RRT) and its variants have been widely used in robot motion planning due to their fast search in state space. However, they perform not well in many complex environments since the motion planning needs to simultaneously consider the geometry constraints and differential constraints. In this article, we propose a novel robot motion planning algorithm that utilizes multi-tree to guide the exploration and exploitation. The proposed algorithm maintains more than two trees to search the state space at first. Each tree will explore the local environment. The tree starts from the root will gradually collect information from other trees and grow towards the goal state. This simultaneous exploration and exploitation method can quickly find a feasible trajectory. We compare the proposed algorithm with other popular motion planning algorithms. The experiment results demonstrate that our algorithm achieves the best performance on different evaluation metrics. 

\end{abstract}

\section{INTRODUCTION}
Designing a feasible trajectory that satisfies global obstacles constraints for a robot from the start state to the goal state is the main underlying idea of robot motion planning \cite{robotmotion}. Over the past few decades, the motion planning problem has been widely studied \cite{studyofmootion}, and many well-known motion planning algorithms have been proposed. The graph-based planning algorithms, such as the A* \cite{A*} and Dijkstra’s algorithm \cite{Dijkstra} are used to find an optimal trajectory through graphs. However, the above methods all rely on the explicit representation of obstacles, which may lead to an enormous computational burden in high-dimensional environments \cite{computation}. Therefore, in order to avoid the explicit construction of obstacles, the sampling-based methods \cite{sbp} are proposed. Sampling-based methods random sample states from the whole space and perform collision detection to build a graph or tree-like structure. This structure can be used to find a feasible trajectory that leads the robot to the desired state.

The sampling-based planning algorithms can be broadly classified into single-query ones and multi-query ones. The classic algorithm of single-query is Rapidly-exploring Random Tree (RRT) \cite{RRT}, that aims to find a feasible solution from the start state to the goal state as quickly as possible. Its search process is similar to the growing and spreading of a tree. However, in high-dimensional environments, the outward expansion of single-query algorithms will be limited. The classic algorithm of multi-query is Probabilistic Road Map (PRM) \cite{PRM}, which first uses random sampling to build a topological graph in the environment, and then searches for the feasible solution on the constructed topological graph through the graph-based planning algorithm. PRM has two phases: the learning and query phases, which means that it is not an anytime algorithm \cite{anytime}. The query phase only starts after the learning phase is completed. Both RRT and PRM require much time to find a feasible solution in complex environments. Searching with multiple trees is an effective time-saving planning method. Multidirectional search maintains more than two trees to search the state space: a rooted tree from the start state and other trees from a random state in the state space. Each tree will explore the local environment. When any two of the multiple trees are close enough, they are joined by solving a two-point Boundary Value Problem (BVP). However, two-point BVP is non-trivial to be solved. There are some common two-point BVP solvers, such as the shooting method \cite{shooting} and linearization method \cite{linear}, but they are time-consuming, and sometimes they do not return reasonable results.

In order to solve the above problems, Multi-Tree (MT-RRT), a novel algorithm based on RRT to improve the planning performance and avoid using two-point BVP solvers, is proposed in this article. MT-RRT still maintains more than two trees: a rooted tree and other random trees. When the rooted tree is close enough to a random tree in the space, the nodes of this random tree will be modeled by Gaussian mixture model (GMM) to serve as the heuristic information to guide the growth of the rooted tree. By heuristic search, the solution of the two-point BVP is avoided, and the rooted tree will gradually collect information from these random trees to quickly grow towards the goal state. MT-RRT algorithm combines the tree's connectivity from RRT and the high visibility from PRM. This simultaneous exploration and exploitation method makes MT-RRT utilize sampling states with more visibility. Thus the connection efficiency of sampling states is improved.

The rest of this article is organized as follows. The Section II first introduces related work on RRT-based motion planning and multi-query search strategy. Formulation of motion planning and related functions are presented in Section III, and the details of the MT-RRT algorithm are explained in Section IV. The results and analysis of simulation experiments are reported in Section V, and finally, we draw conclusions in Section VI.

\section{Related Work}
There are many variants and improvements of RRT in previous research for robot motion planning. Kuffner and LaValle \cite{rrt-connect} proposed the RRT-Connect algorithm, which uses two trees for bidirectional search, and adds a greedy strategy based on the growth method of the RRT to reduce useless searches. Gammell \emph{et al}. \cite{Informed_rrt} proposed Informed-RRT*, which is mainly to accelerate convergence to an optimal solution by confining the sampling space to an elliptical region, and gradually shrinks the elliptical region as the trajectory length decreases. But the biased sampling of Informed-RRT* only activates after a feasible solution is found. Some researches are based on the heuristic biasing to guide the whole search process, such as using heuristic biasing based on costs discovered by exploring the space to guide RRT growth \cite{heuristicbias}, growing towards to the goal region with A* heuristic biased sampling \cite{sample-based-A*}, augmenting RRT-planners with growing multiple local trees \cite{localtree}, and proposing the importance of balancing exploration and exploitation in planning \cite{balance}. 

Recently, methods to explore with multiple trees have also been proposed. Ghosh \emph{et al}. \cite{kb-rrt} proposed a KB-RRT algorithm, which combines bidirectional RRT and kinematic constraints for the robot motion planning in a cluttered environment. Wang \emph{et al}. \cite{b2u} proposed a B2U-RRT algorithm, which maintains two trees: one from the start state and the other from the goal state. When these two trees are close enough, the tree from the goal state acts as a heuristic, guiding the tree from the start state. A learning-based multi-RRTs (LM-RRT) \cite{lm-rrt} approach was proposed to address the robot motion planning in narrow space. However, if space is limited with no apparent narrow passages, such as complex mazes and clutters, the growth of these planners will be bottlenecked. Because the location of the tree is hindered by the environment, it may be challenging to choose a suitable tree to add sampling states. Lai \emph{et al}. \cite{rrdt} proposed the RRdT algorithm, which uses incremental multiple disjointed trees to exploit state space. RRdT achieves high sampling efficiency via multi-query and the balance of exploration and exploitation even in a complex environment. However, it is path planning without considering the geometry and differential constraints.

The proposed MT-RRT in this paper can complement above limitations. For one thing, MT-RRT uses incremental multiple trees to exploit, which makes it obtain high visibility even in tightly constrained space, and for another MT-RRT continuously utilizes information from heuristic trees to quickly find the goal state without requiring a two-point BVP solver.

\section{PRELIMINARY}
This section focused on some definitions and notations. The statement of robot motion planning, RRT extend function, and the cost function are separately introduced in Sections III-A, III-B, and III-C.

\subsection{Robot Motion planning Formulation}
Let $\mathcal X$ denote the state space. Let $\mathcal X_{obs} \in \mathcal X$ denote the obstacle space, and obstacle-free space can be $\mathcal X_{free} = \mathcal X \backslash \mathcal X_{obstacle}$. The objective of the robot motion planning is to find a feasible trajectory with motion control (the geometry constraints and differential constraints). 

In robot motion planning, let $\mathcal{U} \in \mathbb{R}^{n}$ denote the control space, and the planned trajectory can be considered to find a series of motion controls $u:[0,T]\rightarrow \mathcal U$. Therefore, the dynamics equation of robot motion control \cite{motioncontrol} can be formulated as
\begin{equation}
    {x}(t+1) = f(x(t), u(t))
\end{equation}
where $u(t) \in \mathcal {U}$, and the ${x}(t+1)$ and the ${x(t)}$ represent the two adjacent states.

With motion control, find a trajectory from a start state $\mathcal X_{start}$ to a goal state $\mathcal X_{goal}$ in state space $\mathcal X_{free}$. The trajectory is said to be feasible if $x : [0, T] \rightarrow \mathcal X_{free}$, $x(0) = \mathcal X_{start}$, $x(T) = \mathcal X_{goal}$, and for $\forall t \in [0, T]$, $x(t) \in \mathcal X_{free}$. 




\subsection{RRT Algorithm}

\begin{algorithm}  
  \caption{RRT}  
  \KwIn{ $\mathcal X_{start}$, $\mathcal X_{goal}$, $\mathcal G_{map}$}  
  \KwOut{$\tau_{tree}$}  
  $\mathcal V \gets \{\mathcal X_{start}\}, \mathcal E \gets \emptyset, \tau_{tree} \gets (\mathcal N, \mathcal E)$\; 
\For{$i=1$; $i<n$; $i++$ }
{
     $x_{rand} \gets RandomState (\mathcal G_{map}) $\;
     $x_{new} \gets Extend(\tau_{tree},x_{rand})$\;
    \If {$x_{new} \in Region(\mathcal X_{goal})$}
    {
        \Return $\tau_{tree}$\;
    }
} 
\end{algorithm} 

RRT is an efficient planning method in multidimensional space, and Algorithm 1 shows the workflow of classical RRT algorithm. RRT uses a start state $\mathcal X_{start}$ as the root and extends a random branch $x_{new}$ by random sampling state $x_{rand}$ in the whole state space $\mathcal G_{map}$. When the tree $\tau_{tree}$ contains the goal state $\mathcal X_{goal}$, a valid trajectory can be found from $\mathcal X_{start}$ to $\mathcal X_{goal}$.

\subsection{Extend and Cost Function}

\begin{algorithm}  
  \caption{Extend Function}  
  \SetKwFunction{Extend}{Extend}
  \SetKwProg{Fn}{Function}{:}{}
    \Fn {\Extend {$\tau_{tree}, x_{rand}$}}
    {
        $x_{near} \gets NearestNeighbor(\tau_{tree},x_{rand})$\;
        $\upsilon_{current}, \omega_{current} \gets x_{near}$\;
        $\upsilon_{max}, \upsilon_{min} \gets TimeStep(\upsilon_{current}, \alpha_{\upsilon})$\;
        $\omega_{max}, \omega_{min} \gets TimeStep(\omega_{current}, \alpha_{\omega})$\;
        $\delta_\upsilon \gets \Delta(\upsilon_{max}, \upsilon_{min}, \delta_{n\upsilon})$\;
        $\delta_\omega \gets \Delta(\omega_{max}, \omega_{min}, \delta_{n\omega})$\;
        $\mathcal C \gets \emptyset$\;
        \For{$i=0$; $i<=\delta_{n\upsilon}$; $i++$ }
            {
            \For{$j=0$; $j<=\delta_{n\omega}$; $j++$ }
                {
                 $x_{new} \gets Control(\delta_\upsilon, \delta_\omega, i, j)$\;
                 $C_{value} \gets Cost(x_{new}, x_{rand})$\;
                 $\mathcal C \gets \{ C_{value}\}$\;
                    \If {$C_{value} == \arg \min(\mathcal C)$}
                    {
                        \If{$ObstacleFree(x_{new}, x_{near})$}
                            {
                             $\mathcal V \gets AddNode(x_{new})$\;
                             $\mathcal E \gets AddEdge(x_{new}, x_{near})$\;
                             \Return $x_{new}$\;
                            }
                    }
                }
            }
    }

\end{algorithm} 

\begin{algorithm}  
  \caption{Cost Function}  
  \SetKwFunction{Cost}{Cost}
  \SetKwProg{Fn}{Function}{:}{}
    \Fn {\Cost {$x_{new}, x_{rand}$}}
    {
    $Cost_{distance} \gets \frac{||x_{new} - x_{rand}||}{||\mathcal X_{start} - \mathcal X_{goal}||}$\;
    $Cost_{angle} \gets \arctan(\frac{x^v_{new} - x^v_{rand}}{x^h_{new} - x^h_{rand}}) - x^\theta_{new}$\;
    $C_{value} \gets w_1*Cost_{distance} + w_2*\lvert Cost_{angle} \rvert$\;
    \Return $C_{value}$\;
    }
\end{algorithm}    

Algorithm 2 and Algorithm 3 define the extend and cost function of the RRT algorithm, which is used in robot motion planning. In extend function, when we get a random sampling state $x_{rand}$, we can find the nearest state $x_{near}$ according to the distance metric. If we know the current velocity $\upsilon_{current}$, acceleration $\alpha_\upsilon$, angular velocity $\omega_{current}$, and angular acceleration $\alpha_\omega$, we can get the range of velocity $[\upsilon_{min}, \upsilon_{max}]$ and the range of angular velocity $[\omega_{min},\omega_{max}]$ at a particular time interval $TimeStep$. Therefore, the range of velocity and angular velocity can be discretized according to their discrete fractions $\delta_{n\upsilon}$ and $\delta_{n\omega}$. According to different combinations of linear and angular velocities, some candidate states $x_{new}$ can be obtained. When the cost value $C_{value}$ calculated by the Cost Function is minimal, the corresponding candidate state will be considered to connect to the tree $\tau_{tree}$. If this connection is collision-free, the node $x_{new}$ and the edge $(x_{new}, x_{near})$ will be added to the tree $\tau_{tree}$. If not, this expansion will be considered invalid. This expansion process will continue until the new expansion state reaches the goal or iteration limit is reached.

This cost function \cite{chi2018risk} calculates the deviation of the distance and angle between two states. The smaller the cost value, the less time the robot spends traveling between the two states. The details of this cost function are shown in Algorithm 3, where $Cost_{distance}$ and $Cost_{angle}$ represent the cost of distance and angle, respectively. $w_1$ and $w_2$ are parameters that adjust the balance between distance and angle. In this article, $w_1$ and $w_2$ take 1 and 0.3, respectively. $x_{new}$ is the state of the candidate point, and $x_{rand}$ is the state of the sampling point. $||\cdot||$ is the Euclidean norm. The superscripts $h$, $v$, and $\theta$ are the state's horizontal coordinate, vertical coordinate, and angle value.

\section{MT-RRT algorithm}
In this section, we describe the details of the proposed MT-RRT algorithm.

\begin{algorithm}  
  \caption{MT-RRT algorithm}  
  \KwIn{$\mathcal X_{start}$,$\mathcal X_{goal}$,$\mathcal G_{map}$}  
  \KwOut{ $\tau_{root}$}  

$\mathcal V_{root} \gets \{\mathcal X_{start}\}, \mathcal E_{root} \gets \emptyset,$
$\tau_{root} \gets (\mathcal N_{root}, \mathcal E_{root})$\; 

$ \mathcal V_{info} \gets \{node_i\}, \mathcal E_{info} \gets \emptyset,$
$\tau_{info} \gets (\mathcal V_{info}, \mathcal E_{info}),$
$\mathcal T \{\tau_{info}\} \gets RandomGenerate(\mathcal G_{map})$\;

\While {$n \leq N$}
{
    \If{$C\_Tree(\tau_{root}, \mathcal T \{\tau_{info}\}) == FALSE$}
    {
        $x_{rand} \gets RandomState(\mathcal G_{map})$\;
        \If {$Dist(x_{rand}, \mathcal \tau_{root}) < \lambda$}
        {
            $x_{new} \gets Extend(\tau_{root},x_{rand})$\;
        }
        \ElseIf {$Dist(x_{rand}, \mathcal T\{\mathcal \tau_{info}\}) < \lambda$}
        {
            \If{$ObstacleFree(x_{rand}, \mathcal T\{\mathcal \tau_{info}\}$}
            {
            $\mathcal T\{\tau_{info}\} \gets AddNode(x_{rand})$\;
            }
        }
        \Else
        {
            $\tau_{info} \gets RandomGenerate(\mathcal G_{map})$\;
            $\mathcal T\{\tau_{info}\} \gets AddTree(\tau_{info})$\;
        }
    }
    \If{$C\_Tree(\tau_{root}, \mathcal T\{\tau_{info}\}) == TRUE$}
    {
        $\tau_{tree1},\tau_{tree2} \gets C\_Tree(\tau_{root}, \mathcal T\{\tau_{info}\})$\;
        $node_1,node_2 \gets C\_Tree(\tau_{root}, \mathcal T\{\tau_{info}\})$\;
            \If{$\tau_{tree1} ==\tau_{root}$}
            {
                $x_{rand} \gets HeuristicState(\tau_{tree2})$\;
                $x_{new} \gets Extend(\tau_{root}, x_{rand})$\;
                $Delete(\tau_{tree2})$\; 
                    \If {$x_{new} \in Regin(\mathcal X_{goal})$}
                     {
                        \Return $\tau_{root}$\;
                     }
            }
            \Else
            {
            \If{$ObstacleFree(node_1,node_2)$}
                {
                $ConnectedNode(node_1,node_2)$\;
                }
            $\tau_{tree1} \gets ExtendTree(\tau_{tree1},\tau_{tree2})$\;
            $Delete(\tau_{tree2})$\;
            }
    }
}
\end{algorithm}

\begin{figure*}
\centering
    \subfigure[Step 1.]{\includegraphics[width=0.5\columnwidth]{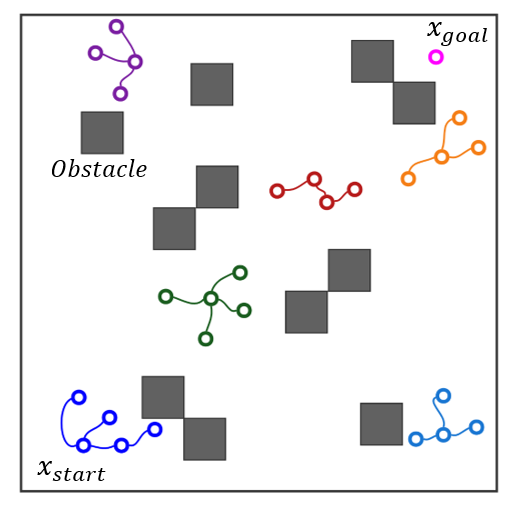}}
    \subfigure[Step 2.]{\includegraphics[width=0.5\columnwidth]{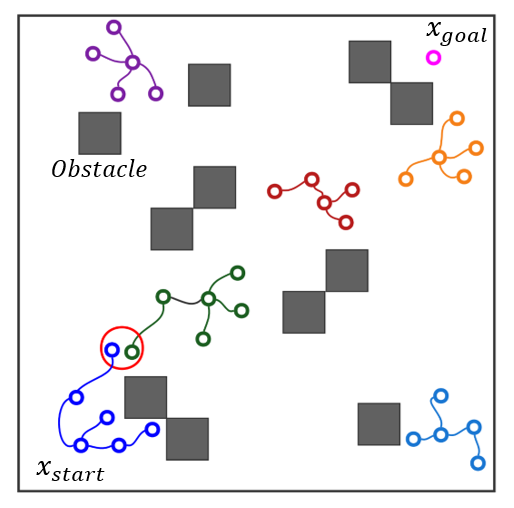}}
    \subfigure[Step 3.]{\includegraphics[width=0.5\columnwidth]{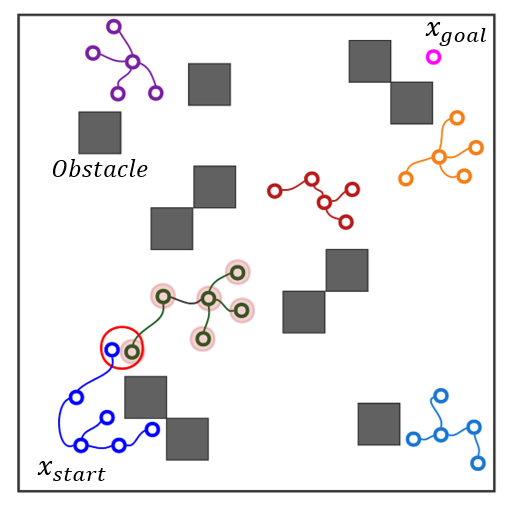}}
    \subfigure[Step 4.]{\includegraphics[width=0.5\columnwidth]{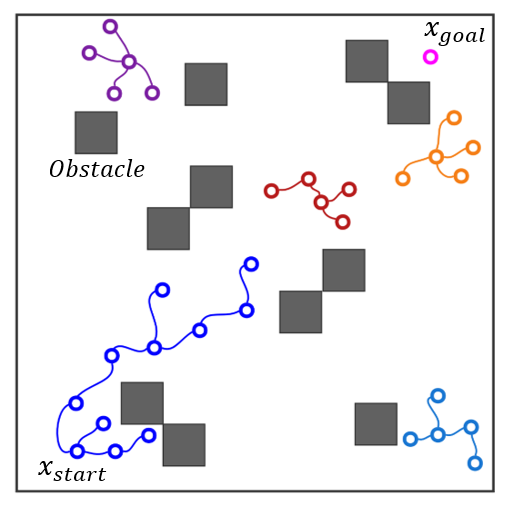}}
\caption{The planning process of MT-RRT. The dark blue circle indicate the start state, and the rose red circle indicate the goal state. Blank area represents free space, and the rectangles represent obstacles. (a) The rooted tree $\tau_{root}$ and heuristic trees $\mathcal T\{\tau_{info}\}$ begin to grow. (b) The heuristic tree $\tau_{info}$ (one of $\mathcal T\{\tau_{info}\}$) is close enough to $\tau_{root}$. (c) The nodes of $\tau_{info}$ are modeled with GMM to serve as heuristic information. (d) $\tau_{root}$ is growing by using this heuristic information of $\tau_{info}$, and then $\tau_{info}$ is removed from the space.}
\label{step}
\end{figure*}

The intuitive idea of MT-RRT is to utilize heuristic trees to guide the rooted tree to extend towards the goal region. This simultaneous exploration and exploitation method can quickly find a feasible trajectory.

Different from the unidirectional and bidirectional RRT, MT-RRT maintains more than two trees searching in the state space: the rooted tree $\tau_{root}$ from the start state $\mathcal X_{start}$ and heuristic trees $\mathcal T\{\tau_{info}\}$ are random generated through $RandomGenerate$ in the state space $\mathcal G_{map}$. Each tree will explore the local environment. When $\tau_{root}$ is close enough to $\tau_{info}$, which is one of  $\mathcal T\{\tau_{info}\}$, $\tau_{info}$ will serve as the heuristic to guide $\tau_{root}$. $\tau_{root}$ gradually collects information from $\mathcal T\{\tau_{info}\}$ and quickly grows towards the goal state.

\emph{1)Connect-Nodes Stage, Line 4-13, Algorithm 4:} In this stage, the node $x_{rand}$ from $RandomState$ will be added to $\tau_{root}$ or any tree of $\mathcal T\{\tau_{info}\}$. If the distance $Dist(x_{rand},\tau_{root})$ between $x_{rand}$ and $\tau_{root}$ is less than $\lambda$, $x_{rand}$ will be added to $\tau_{root}$ through $Extend$ function; if the distance $Dist(x_{rand},\mathcal T\{\tau_{info}\})$ between $x_{rand}$ and $\mathcal T\{\tau_{info}\}$ is less than $\lambda$, $x_{rand}$ will be added to a corresponding tree of $\mathcal T\{\tau_{info}\}$ through $AddNode$; otherwise, a new tree $\tau_{info}$ will be generated at $x_{rand}$ through $RandomGenerate$ and this new tree $\tau_{info}$ will be added to $\mathcal T\{\tau_{info}\}$ through $AddTree$. The result of this stage is as shown in Fig. \ref{step}(a).

\emph{2)Connect-Trees Stage, Line 14-27, Algorithm 4:} In this stage, if two trees are close enough to each other (shown in Fig. \ref{step}(b)), they will be merged into one tree. We can get the two trees ($\tau_{tree1},\tau_{tree2}$) and the two adjacent nodes ($node_1, node_2$) through the function of connection trees ($C\_Tree$). If one of these two trees ($\tau_{tree1}$) is $\tau_{root}$, the other tree ($\tau_{tree2}$) will be served as the heuristic to guide $\tau_{root}$ to get biased sampling node $x_{rand}$ through $HeuristicState$. Then the new node $x_{new}$ can be extended to the $\tau_{root}$ through $Extend$ function as shown in Fig. \ref{step}(d), and $\tau_{tree2}$ will be removed. Otherwise, the two adjacent nodes ($node_1,node_2$) will be connected, and all nodes from one tree ($\tau_{tree2}$) will be added to the other tree ($\tau_{tree1}$).

We can construct a Gaussian distribution at each node on the heuristic tree (shown in Fig. \ref{step}(c)), then the entire heuristic tree can be combined by the Gaussian mixture model (GMM). With the GMM, the planner can achieve biased sampling guided by the heuristic tree. The GMM can be defined as
\begin{equation}
p(x)\!=\!\sum_{j=1}^{\kappa} \frac{1}{\kappa} \frac{1}{\sqrt{2 \pi \Sigma_{j}^{2}}} \exp \left(\!-\!\frac{\left(x\!-\!\mu_{j}\right)^{T} \Sigma_{j}^{-1}\left(x\!-\!\mu_{j}\right)}{2}\right)
\end{equation}
where $p(x)$ is the probability density of the $x$ state, $\mu$ and $\sigma$ refer to the corresponding mean and standard deviation, and $\kappa$ denotes the number of nodes selected from the heuristic tree. In 2D,
\begin{equation}
\mu=\left[x_{x}, x_{y}\right], \quad \Sigma=\left[\begin{array}{cc}
\sigma_{x} & 0 \\
0 & \sigma_{y}
\end{array}\right]
\end{equation}

In the heuristic sampling process, the heuristic tree will be deleted after the heuristic information is provided, which ensure that the heuristic tree will continue to generate in the state space to provide new exploration and exploitation information. At the same time, when there is no heuristic information, the rooted tree will also random sample in the state space. Therefore, the robot in a complex environment can achieve fast forward growth state before the heuristic tree is found. In this way, a valid trajectory from a start state to a goal state can be quickly found and the probabilistic completeness \cite{completeness} can be ensured.

\section{EXPERIMENT}

\begin{figure*}
\centering
    \subfigure[Room, RRT.]{\includegraphics[width=0.55\columnwidth]{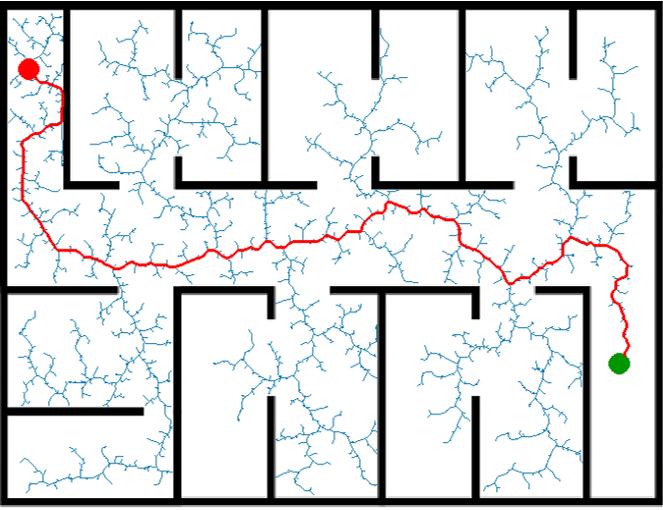}}
    \hspace{0.5cm}
    \subfigure[Clutter, RRT.]{\includegraphics[width=0.55\columnwidth]{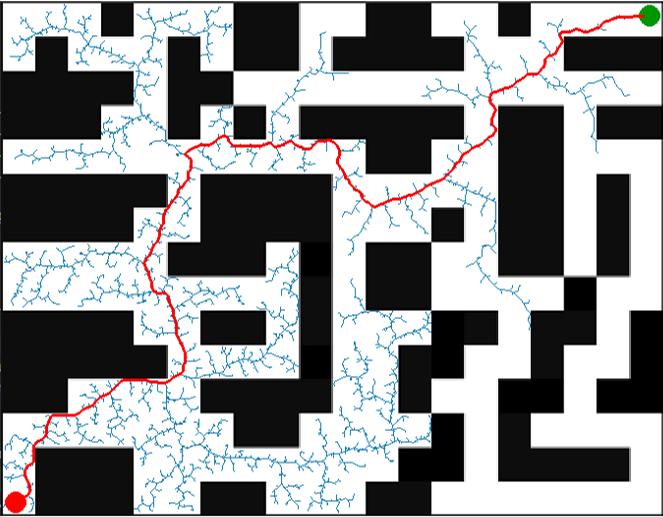}}
    \hspace{0.5cm}
    \subfigure[Maze, RRT.]{\includegraphics[width=0.55\columnwidth]{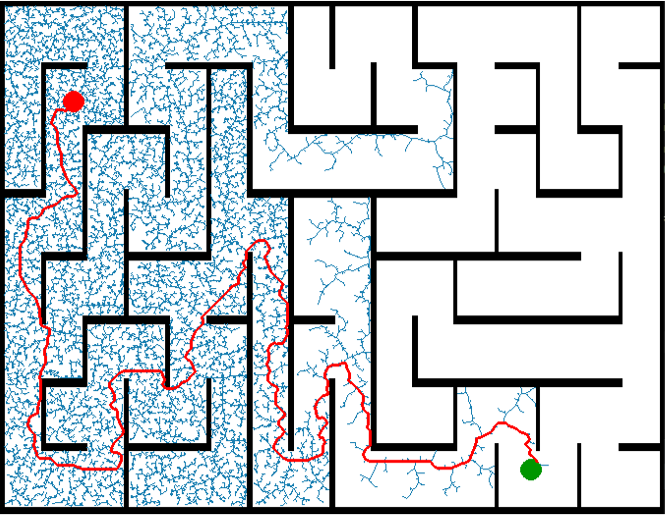}}
    \hspace{0.5cm}
    \subfigure[Room, B2U-RRT.]{\includegraphics[width=0.55\columnwidth]{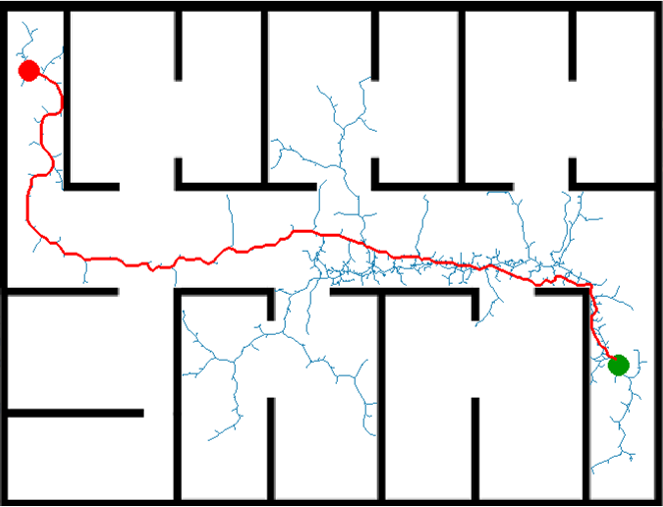}}
    \hspace{0.5cm}
    \subfigure[Clutter, B2U-RRT.]{\includegraphics[width=0.55\columnwidth]{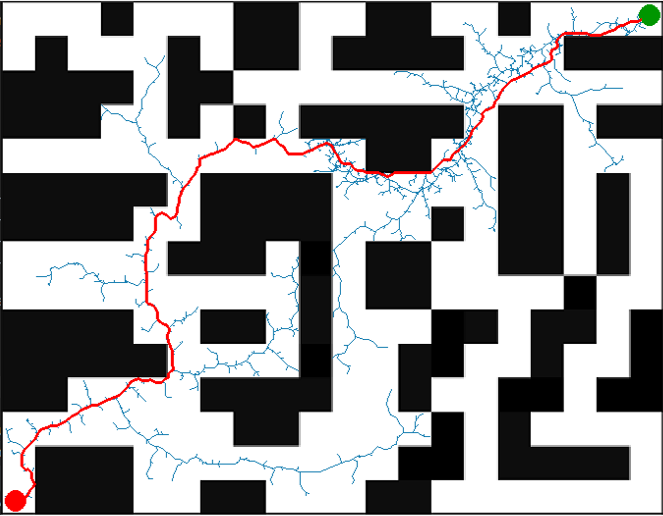}}
    \hspace{0.5cm}
    \subfigure[Maze, B2U-RRT.]{\includegraphics[width=0.55\columnwidth]{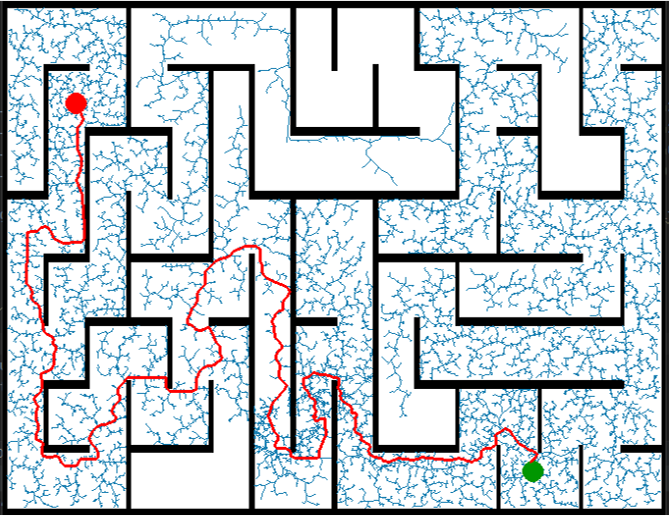}}
    \hspace{0.5cm}
    \subfigure[Room, MT-RRT.]{\includegraphics[width=0.55\columnwidth]{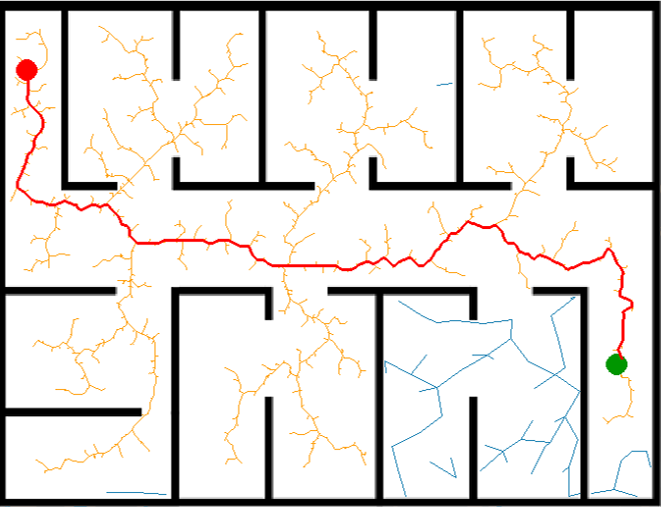}}
    \hspace{0.5cm}
    \subfigure[Clutter, MT-RRT.]{\includegraphics[width=0.55\columnwidth]{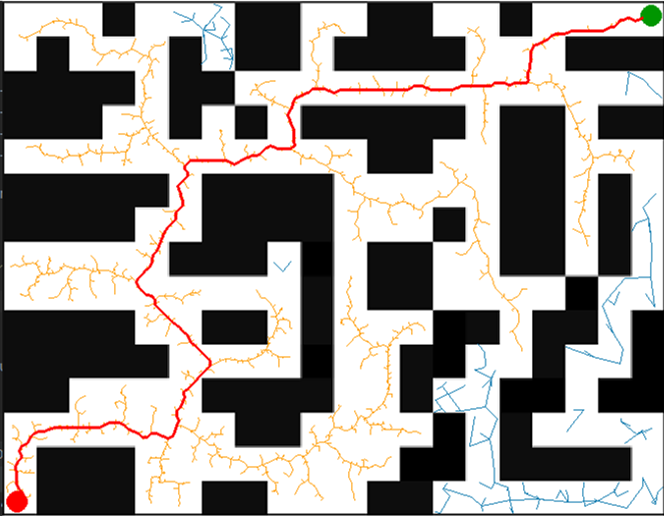}}
    \hspace{0.5cm}
    \subfigure[Maze, MT-RRT.]{\includegraphics[width=0.55\columnwidth]{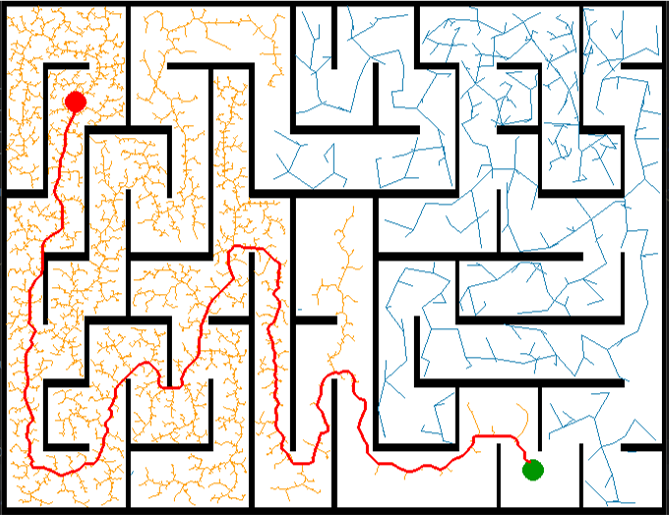}}
\caption{Illustration of three simulation environment. The red point represents the start state and the green point represents goal state. Blank area represents free space and black region represents obstacles. The red curve represents the initial trajectory found by the planner. In the RRT and B2U-RRT, the blue branches denote the searched tree. Close to the goal region of B2U-RRT, the dense blue branches represent the traces of the heuristic search. In the MT-RRT, the orange branches denote the tree growing from the start state and the blue branches denote trees growing from other states to provide heuristic information.}
\label{env}
\end{figure*}

\begin{figure*}
\centering
    \subfigure[Room, Mean.]{\includegraphics[width=0.66\columnwidth]{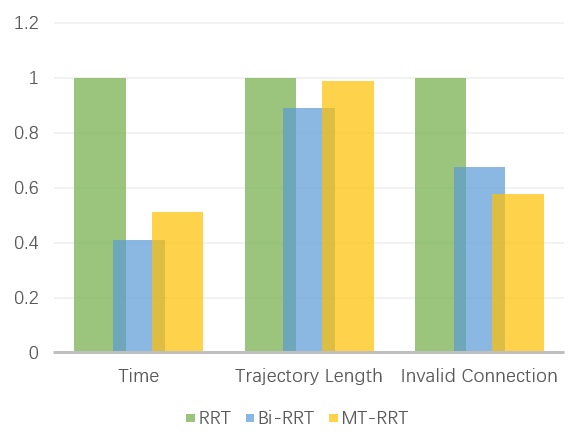}}
    \subfigure[Clutter, Mean.]{\includegraphics[width=0.66\columnwidth]{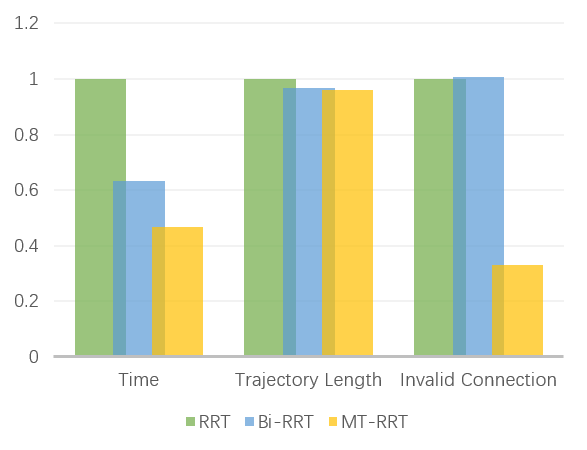}}
    \subfigure[Maze, Mean.]{\includegraphics[width=0.66\columnwidth]{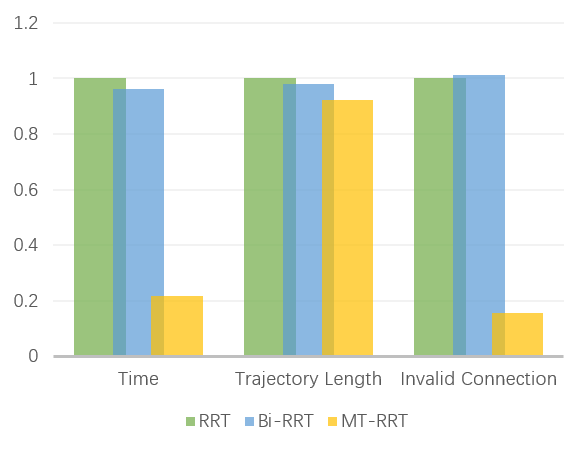}}
    \subfigure[Room, Variance.]{\includegraphics[width=0.66\columnwidth]{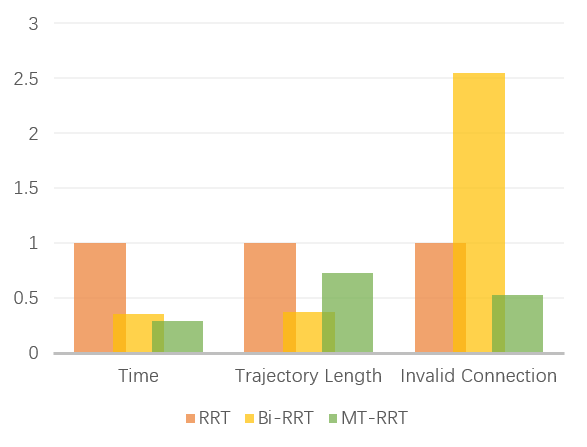}}
    \subfigure[Clutter, Variance.]{\includegraphics[width=0.66\columnwidth]{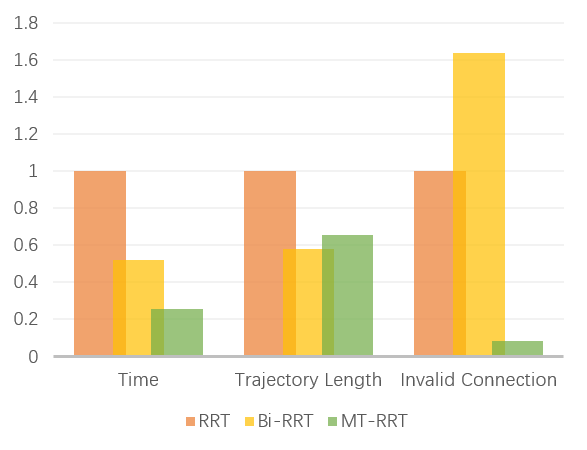}}
    \subfigure[Maze, Variance.]{\includegraphics[width=0.66\columnwidth]{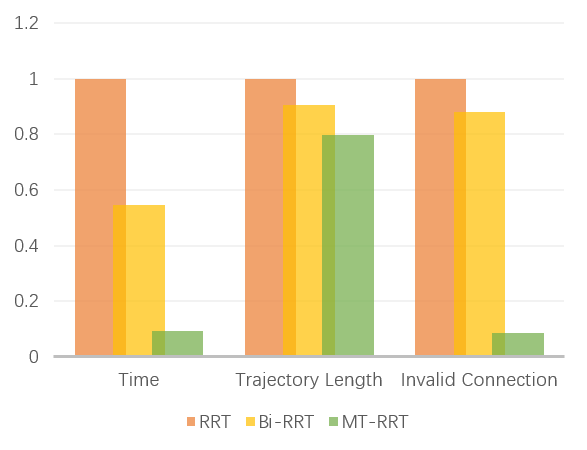}}

\caption{Illustration of experimental statistics. The height of the histogram denotes the value of mean or variance. All results are normalized to the corresponding value of RRT.}
\label{data}

\end{figure*}
In this section, we compare two popular motion planning algorithms, including the RRT and the B2U-RRT \cite{b2u}, with the proposed MT-RRT algorithm. Both RRT and B2U-RRT use the extension function of Algorithm 2, as we introduced in Section III. Instead of solving two-point BVP, B2U-RRT uses a heuristic search (using the tree extending from the goal state as heuristic information) when the two trees are close enough. We evaluate the performance of RRT, B2U-RRT, and MT-RRT in three simulation environments, which are shown in Fig. \ref{env} with increasing complexity from left to right. The left column shows a room environment where the robot needs to go through a few turns to find the goal state. The environment in the middle column contains a lot of cluttered block obstacles, and the robot needs to pass through these obstacles to reach the goal state. The right one is a maze environment with many turns and dead ends. The solution to the goal can only be found after successfully going through numerous turns. We used three metrics ($Time$, $Trajectory$ $Length$ and $Invalid$ $Connection$) to evaluate the performance of RRT, B2U-RRT and MT-RRT. The meanings of the three metrics are: $Time$ means the time cost of finding the initial solution, $Trajectory$ $Length$ is the length of trajectory from start to goal, and $Invalid$ $Connection$ is the number of invalid samples due to intermediate connections being invalid.

The map size for these three environments is 450 $\times$ 350 pixels. RRT, B2U-RRT and MT-RRT were implemented by python with the same planning framework and tested on Intel i5-10400 CPU with 32GB RAM. Each algorithm is executed 50 times in each environment to get reasonable statistics.

Taking the mean and variance of 50 experiments we can get the experimental results in Fig. \ref{data}, where (a)(d), (b)(e), and (c)(f) are the mean and variance of RRT, B2U-RRT and MT-RRT in a room, clutter, and maze environments, respectively. Due to the significant differences in the values of these three metrics, we standardized the results based on the value of RRT. Therefore, the value of each metric of RRT in Fig. \ref{data} is 1. 

According to the mean of each metric in the three environments, we find that MT-RRT is slightly inferior to B2U-RRT in the simple room environment, but still better than RRT. However, as the complexity of the environment increases, like the clutter and maze, MT-RRT significantly outperforms RRT and B2U-RRT, especially in the $Time$ of finding an initial solution. Compared with RRT and B2U-RRT, MT-RRT has the least number of failed connection nodes in all three environments, which shows MT-RRT has high sampling efficiency in both simple and complex environments. Besides, in these three environments, B2U-RRT consistently outperforms RRT in the time to get solution, illustrating the benefits of multiple trees in planning.

Comparing the variance of each metric in the three environments, MT-RRT shows robust performance. B2U-RRT fluctuates greatly in connectivity because B2U-RRT's heuristic information comes from the tree growing from the goal state, and the expansion of this tree has a great influence on B2U-RRT's heuristic search. However, MT-RRT is very stable on connections. Because MT-RRT's heuristic information comes from growing trees that are continuously updated. The result of variance in $Valid$ $Connection$ shows that MT-RRT is more robust in heuristic search than B2U-RRT.

\section{CONCLUSIONS}
In this article, the proposed MT-RRT is a sampling-based motion planner with incremental multi-tree. MT-RRT utilizes heuristic information from the trees that are continuously updated and growing in state space. Due to this way of simultaneous exploration and exploitation, MT-RRT can quickly guide a robot towards the goal. Experimental results demonstrate that our proposed algorithm achieves excellent performance in different metrics compared with RRT and B2U-RRT. With our proposed method, both the efficiency and robustness of robot motion planning in finding an initial feasible solution have been substantially improved. Besides, social and psychological constraints can also be considered in MT-RRT to realize robot motion planning that obeys social norms \cite{social}.



\bibliographystyle{IEEEtran} 
\bibliography{refs} 

\end{document}